\documentclass[conference]{ieeeconf}

\usepackage[hidelinks]{hyperref}
\usepackage{url}
\usepackage{amsmath,amssymb,amsfonts}
\usepackage{cite}
\usepackage{algorithmic}
\usepackage{graphicx}
\usepackage{booktabs}
\usepackage{multirow}   
\usepackage{siunitx}

\usepackage{xcolor}
\usepackage{paralist}
\usepackage{listings}
\usepackage{orcidlink}
\usepackage{tabularx}
\usepackage{cleveref}
\crefname{section}{Sect.}{Sect.}
\Crefname{section}{Sect.}{Sect.}
\crefname{listing}{Lst.}{Lst.}
\Crefname{listing}{Lst.}{Lst.}
\crefname{figure}{Fig.}{Fig.}
\Crefname{figure}{Fig.}{Fig.}

\usepackage{dirtree}

\newcommand{\Fone}{\textsf{\textbf{F1}}}
\newcommand{\Ftwo}{\textsf{\textbf{F2}}}
\newcommand{\Fthree}{\textsf{\textbf{F3}}}
\newcommand{\Ffour}{\textsf{\textbf{F4}}}

\newcommand{\Aone}{\textsf{\textbf{A1}}}
\newcommand{\Aoneone}{\textsf{\textbf{A1.1}}}
\newcommand{\Aonetwo}{\textsf{\textbf{A1.2}}}
\newcommand{\Atwo}{\textsf{\textbf{A2}}}

\newcommand{\Ione}{\textsf{\textbf{I1}}}
\newcommand{\Itwo}{\textsf{\textbf{I2}}}
\newcommand{\Ithree}{\textsf{\textbf{I3}}}

\newcommand{\Rone}{\textsf{\textbf{R1}}}
\newcommand{\Roneone}{\textsf{\textbf{R1.1}}}
\newcommand{\Ronetwo}{\textsf{\textbf{R1.2}}}
\newcommand{\Ronethree}{\textsf{\textbf{R1.3}}}

\newcommand{\tl}[1]{%
  \begingroup
  \def\c{#1}%
  \ifx\c\empty\texttt{[\,]}\else
    \textcolor{#1}{\Large$\bullet$}%
  \fi
  \endgroup
}

\definecolor{codegreen}{rgb}{0,0.6,0}
\definecolor{codegray}{rgb}{0.5,0.5,0.5}
\definecolor{codepurple}{rgb}{0.58,0,0.82}

\lstdefinestyle{mystyle}{
    backgroundcolor=\color{white},   
    commentstyle=\color{codegreen},
    keywordstyle=\color{codegreen},
    numberstyle=\tiny\color{codegray},
    stringstyle=\color{codepurple},
    basicstyle=\ttfamily\scriptsize,
    breakatwhitespace=false,         
    breaklines=true,                 
    captionpos=b,                    
    keepspaces=true,                 
    numbers=none,                    
    numbersep=5pt,                  
    showspaces=false,                
    showstringspaces=false,
    showtabs=false,                  
    tabsize=2,
    otherkeywords={BIND, AS, IF, GROUP, SUM, COUNT, DISTINCT, FILTER, GROUP_CONCAT, SEPARATOR, FILTER, NOT, EXISTS}
}

\newcommand{\RoboVAST}{\textsc{RoboVAST}}

\hypersetup{pdfauthor={Argentina Ortega, Samuel Wiest, Frederik Pasch, Nico Hochgeschwender},pdftitle={Replicable Simulation-Based Robot Validation through Provenance}}

\begin{document}
\bstctlcite{IEEEBSTcontrol} %

\title{\textbf{Replicable Simulation-Based Robot Validation through Provenance\\
}}
\author{Argentina Ortega$^{1}\orcidlink{0000-0002-3873-4435}$ \and Samuel Wiest$^{1}\orcidlink{0009-0002-9215-1238}$ \and Frederik Pasch$^{2}\orcidlink{0000-0002-2626-1538}$ \and Nico Hochgeschwender$^{1}\orcidlink{0000-0003-1306-7880}$%
}

\maketitle

\begingroup
  \renewcommand{\thefootnote}{}%
  \footnotetext{$^{1}$ University of Bremen, Germany.
    {\tt\footnotesize \{argentina.ortega, samuel.wiest, nico.hochgeschwender\}@uni-bremen.de}}
  \footnotetext{$^{2}$ Karlsruhe University of Applied Sciences, Germany.
    {\tt\footnotesize frederik.pasch@h-ka.de}}
  \footnotetext{This work has partly been supported by 
    the European Union's Horizon Europe project SOPRANO (Grant No. 101120990).}
\endgroup

\thispagestyle{empty}
\pagestyle{empty}

\begin{abstract}
Robot behavior is often validated through simulation-based testing, yet the replicability of such campaigns depends critically on transparent documentation of how tests are configured, executed, and post-processed. We argue that data provenance, coupled with the FAIR principles (findability, accessibility, interoperability, and reusability), addresses this gap by explicitly tracking links between artifacts and by attaching machine-readable metadata about file origins and key design decisions. Moreover, provenance and metadata cannot be treated as an afterthought confined to final datasets; they must be integrated into the testing processes that generate those datasets so that evidence can be reconstructed end-to-end. We demonstrate this by augmenting an existing simulation-based testing framework with provenance tracking and metadata collection mechanisms, and by using these extensions to enrich a mobile robot navigation dataset with structured provenance and FAIR-aligned metadata. Finally, we discuss obstacles encountered in this integration—such as vocabulary alignment, attribute selection, and adoption of domain standards—and provide actionable recommendations for implementing provenance-centric, FAIR metadata in robotics validation workflows.
\end{abstract}

\section{Introduction}

\noindent 
Deploying robots in real world environments requires thorough validation of its behaviors.
While ongoing research aims to find approaches and tools to address challenges in the field of software testing for robotics~\cite{Araujo2023,Afzal2020},
it remains difficult to replicate results from other studies.
In fact, replicability has been identified as an ongoing issue in robotics~\cite{Bonsignorio2015,leichtmann2022crisis} in general. 
Although datasets are published in academic venues and publications mention supplementary data~\cite{bonsigniorio2017ram}, access to the data itself is constrained by how it is published, described and indexed, and often limited by the longevity of the URLs used, making it difficult to find datasets with specific properties or data types (e.g., types of robots or specific sensors), particularly after some time has passed since its publication.
The FAIR principles---findability, accessibility, interoperability, and reusability---provide concrete criteria for making research artifacts easier to discover, obtain, combine, and reuse across studies~\cite{wilkinson2016fair, Jacobsen2020}. 
Provenance is a key aspect of FAIR and refers to the machine-readable information about the usage, generation and attribution of the elements in a dataset, e.g., which inputs, tools, and processes produced which outputs.
We argue that introducing provenance and adopting FAIR principles in the dataset creation process improves the description of the available data, and thus facilitates the replicability of datasets by making their inputs, configurations, and results easier to find, access, and reuse.

Although studies using simulation-based validation should be easier to replicate than those in real-world environments given our ability to execute the same tasks in the under the same set of conditions, the replicability of these studies is highly dependent on how well the implementation choices (e.g., scenario specifications, software configurations, model transformations, etc.) are documented and understood.
If the path from testing inputs to published results cannot be reconstructed, it is unclear whether an observed effect reflects the robot under study or particularities of the validation setup, a traceability challenge already observed in field testing campaigns~\cite{Ortega2022}. 
Missing or poorly documented data undermines both repeatability (re-running the same campaign with the same input artifacts) and reproducibility (independently re-implementing the campaign based on its documentation). 
We argue that provenance as required by the FAIR principles, i.e., modeled as linked data using well-known vocabularies, can help tackle these traceability issues by encoding the relations between the used entities and the produced results in a machine-readable format, allowing users to more easily exploit this information for repeating, reproducing, and replicating results.

Our paper is structured as follows.
We present related work in~\cref{sec:related-work}. 
In~\cref{sec:case-study} we present our case study:
the creation of a dataset for a mobile robot validation campaign and introduce the FAIR principles in more detail.
In~\cref{sec:modelling} we describe our modeling choices
and show how they describe the artifacts and transformations that constitute the campaign’s validation data. 
To the best of our knowledge, no studies have reported on provenance modeling for robotics data. 
In \cref{sec:implementation}, we demonstrate how we integrated these models into \RoboVAST{}
--- an existing testing framework --- to automate the recording of provenance and metadata.
Finally, in~\cref{sec:dataset} we discuss the resulting dataset and demonstrate how to query the linked data provenance graph. 

In short, our contributions are: 
\begin{inparaenum}[(i)]
\item development of provenance models and metamodels, and machine-readable metadata to meet the FAIR principles in robot validation datasets, 
\item a demonstration of the integration and application of these models into a scenario-based testing framework, and
\item a published \emph{FAIR-by-design} dataset described by provenance concepts captured by the extended framework and queries to the provenance graph to exemplify its use.
\end{inparaenum}

\section{Related Work}
\label{sec:related-work}

\noindent 
Despite promising initiatives to increase the reproducibility in robotics studies~\cite{bonsigniorio2017ram}, 
many experimental robotics papers still do not report enough detail to support reproducibility, including clear evaluation criteria, consistency between methods and criteria, and the information needed to reproduce results~\cite{Faragasso2023}. 
More generally, replicable robotics experimentation has long been understood to require detailed descriptions of assumptions, system parameters, environments, tasks, and benchmarking criteria~\cite{Bonsignorio2015}.
CodeOcean\footnote{\url{https://codeocean.com/explore}} and IEEE Dataport\footnote{\url{https://ieee-dataport.org}} are recommended for Reproducible Articles in the IEEE Robotics and Automation Magazine~\cite{bonsigniorio2017ram}, but both mainly support search over metadata rather than dataset contents and provide only limited filtering. Both also collect basic dataset metadata, although Dataport exposes it only through its web interface, whereas CodeOcean additionally provides a YAML file; Dataport further allows optional documentation uploads, but without guidance on format or presentation.

Even when such information is available, robotic validation campaigns remain difficult to replicate because they depend on physical artifacts that are often unique, locally configured, and sensitive to tolerances, calibration, software versions, and environmental conditions. The difficulty is especially acute in surgical robotics, where reproducibility, replicability, and benchmarking are further constrained by safety, ethical, and intellectual-property concerns~\cite{Faragasso2023}. Recent commentary has therefore called for a stronger shift toward reproducible robotics research, yet the gap between publishing an experimental result and enabling an independently repeatable validation campaign remains substantial~\cite{Bonsignorio2025}.

\looseness-1
Scientific robotic competitions partly address these limitations by defining benchmarking protocols that compare robot performance on well-specified tasks~\cite{Nardi2016}. Many follow established replication principles~\cite{Amigoni2015} and describe competition scenarios, environmental conditions, initialization constraints, evaluation criteria, and observation methods explicitly~\cite{Nguyen2023}. Some competitions also record and share trial data, sometimes with dedicated benchmarking infrastructure~\cite{Thoduka2024,Schneider2015}. Beyond competitions, most robotics datasets, including those in autonomous driving, are curated to measure scientific progress and support new robot capabilities; this is particularly visible in vision-language-action research, which relies on large, heterogeneous demonstration corpora spanning tasks, embodiments, and sensing setups~\cite{Kawaharazuka2025}.

\looseness-1
Machine learning has no universally adopted FAIR-equivalent standard and instead relies on reproducibility checklists, dataset documentation frameworks such as Datasheets for Datasets, model documentation frameworks such as Model Cards, and emerging metadata standards such as Croissant~\cite{Mitchell2019,akhtar2024neurips}. These improve transparency and reuse, but they do not by themselves satisfy stricter FAIR requirements such as persistent identifiers, durable archival access, machine-actionable metadata, formal provenance, interoperable semantics, and long-term stewardship~\cite{wilkinson2016fair,akhtar2024neurips}. The same limitation applies to VLA datasets and models: resources such as Open X-Embodiment and OpenVLA improve accessibility and some interoperability by releasing code, weights, and standardized formats, yet they remain discoverable mainly through papers, project pages, and community knowledge rather than robust archival indexing and fully FAIR-compliant metadata infrastructures~\cite{oneill2024,Kawaharazuka2025,kim25openvla}.

Large-scale surveys of autonomous driving datasets likewise point to the need for standardized formats, labeling guidelines, and access protocols to support interoperability and reuse~\cite{Liu2024}. Access is often shaped by privacy constraints, persistent identification practices vary, and machine-actionable metadata remain uncommon. FAIR principles are more established in disciplines with long traditions of data sharing, such as physics, meteorology, and astronomy, where fewer experimental systems and measurement devices are involved; the first FAIR robotics datasets are emerging where robotics intersects with such disciplines, including marine science and underwater robotics~\cite{motta2023SciData}. In safety- and reliability-oriented validation campaigns, however, data publication remains rare, and reported insights are still mostly descriptive, focusing on which methods, such as simulation-based testing, were used to validate requirements or how standard conformance was achieved~\cite{Afzal2021,Sohail2023}.

\section{Case Study}
\label{sec:case-study}
\noindent
Our motivating use case is the validation of robotics software and how such validation can be made repeatable and reproducible in practice. As an example case study, we created a simulation-based navigation dataset and use it to illustrate how provenance and FAIR-oriented metadata can be layered onto an existing testing campaign and results without changing the core testing logic. This section describes the \textit{baseline} structure of that campaign and dataset, before any explicit provenance or FAIR metadata are introduced.
Concretely, the System Under Test (SUT) is a Turtlebot~4, a mobile robot using the Navigation2 (Nav2) stack in the default \textit{nav2\_bringup} configuration with AMCL for localization, MPPI as the local motion planner, and NavFn as the global path planner~\cite{Macenski2020}. 
Our tests are done in the Gazebo simulator, where the robot executes a single-goal \textit{nav\_to\_pose} mission. 
The test is labeled successful when the final pose is reached within a predefined distance threshold. 
The resulting collection of runs, their configurations, and their logs forms the dataset we aim to apply provenance to.

\subsection{Dataset Design}
\noindent
The dataset centers on a set of models that describe where the robot operates, what task it must perform, and how conditions vary across runs.
Our validation strategy uses scenario-based testing to specify, execute and evaluate our SUT.
An \emph{environment model} captures the physical layout in which the robot navigates; a \emph{scenario file} describes the abstract navigation task within that environment (e.g., navigate to single pose, follow a set of goal poses, etc.); and a \emph{scenario variation file} specifies how scenario parameters, such as start/goal poses or obstacle count, are systematically varied to obtain many instantiated test cases.

The environment is defined by a FloorPlan model~\cite{Ortega2024,Parra2023} that encodes the geometric layout of rooms, corridors, and static obstacles in which the robot navigates. 
From this model, simulation artifacts used for test execution, such as a 3D mesh file and occupancy grid, are generated.  
Environment variations systematically adjust floor plan features (e.g., room dimensions), allowing multiple related map configurations to be tested without redefining the entire environment.

The abstract scenario is specified using the OpenSCENARIO DSL. 
This file formally specifies the abstract test task, including the participating agent (the robot), the robot’s maneuvering sequences, and event execution sequences that describe how the mission is carried out and how additional testing actions, such as data recording, are triggered.
Scenario variability is captured separately in a scenario variation file, which parametrizes both the abstract OpenSCENARIO task and additional parameters relevant to the concrete test configuration. These parameters include the set of start and goal pose(s), the number and placement of obstacles, and the configuration of sensor noise and message dropout. The same mechanism can also vary robot and software configuration files, for example by changing the Nav2 controller or planner settings.
The specifications from these files are then instantiated into concrete test configuration files, each of which fixes all previously variable aspects, such as poses and configuration parameters, into a fully resolved test case. 

\subsection{Dataset Creation}

\noindent The dataset is generated through a systematic simulation-based navigation campaign executed with \RoboVAST\footnote{\url{https://github.com/cps-test-lab/robovast}}, an open-source framework for automated, large-scale integration testing of robotic software in simulation.
\RoboVAST\ builds on the Floorplan-DSL~\cite{Parra2023} for parameterizable indoor environment generation, Scenario Execution~\cite{pasch_scenario_2024} for individual test execution, and Kubernetes-based orchestration to execute many test runs in parallel. 

\begin{figure*}[!htb]
    \centering
    \includegraphics[width=0.9\linewidth]{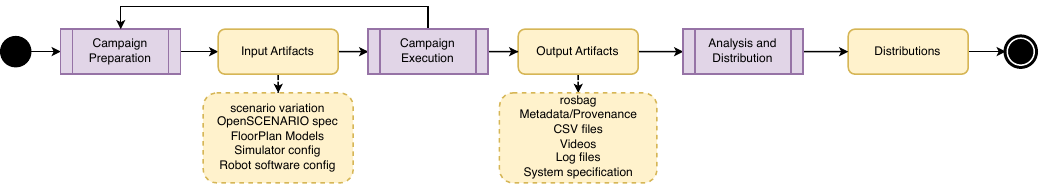}
    \caption{Our dataset creation process} %
    \label{fig:methodology}
    \vspace{-1.25\baselineskip}
\end{figure*}

Our dataset creation follows the three-phase process outlined in ~\cref{fig:methodology}: campaign preparation, campaign execution, and dataset distribution. 
In the preparation phase, we define both \emph{what} to test (SUT, objectives, operating conditions, and success criteria) and \emph{how} to test it (framework, models, and logging). We selected the Turtlebot~4 with Nav2, chose indoor environments and navigation tasks, and configured \RoboVAST\ to execute varied single-goal \textit{nav\_to\_pose} tasks under varying obstacle and sensor-noise conditions. 
During execution, the scenarios are run in simulation to generate raw dataset artifacts, including rosbags, logs, and outcome reports, with runs distinguished by their test configuration files. For analysis, bag files are post-processed into tabular summaries and videos, and all artifacts are packaged with the raw data and logs for publication on a public hosting service such as Zenodo for distribution.

\subsection{Conformance to FAIR Principles}

\noindent
\Cref{tab:gofair-requirements} lists the FAIR principles and the challenges we identified in the iterative dataset creation process.
In the remainder of this paper, we map the identified challenges and FAIR principles to the modeling and implementation mechanisms that we use to address them.

We identify two challenges (mainly) related to modeling the (meta)data of a dataset.
\textbf{C1: Modeling (meta)data and provenance in robotics}. To the best of our knowledge, no studies have reported on the integration of provenance (\Ronetwo) to robotic datasets.
While there are some well-established vocabularies  --- such as PROV\footnote{\url{https://www.w3.org/TR/prov-o/}} for modeling the provenance between elements, DCAT\footnote{\url{https://www.w3.org/TR/vocab-dcat-3/}} to describe dataset and distribution metadata and Dublin Core (DCTERMS)\footnote{\url{http://purl.org/dc/terms/}} to describe metadata such as creators, dates of modification and publication, etc.--- that are suitable for FAIR (meta)data (\Itwo,\Ithree), 
robotics use cases have a wide variety of (meta)data types and properties that are not covered in existing vocabularies, and which are needed for rich descriptions (\Rone). 
\textbf{C2: Standard Metamodels and Controlled Vocabularies}. 
There are few domain-relevant community standards for (meta)data in robotics. Existing metadata descriptions, such as~\cite{ieee7300355}, are not represented in a formal language for knowledge representation (\Ione). To the best of our knowledge, no controlled vocabulary exists to describe dataset subjects or glossary terms.

\setlength{\tabcolsep}{3pt}
\renewcommand{\arraystretch}{0.5}
\newcommand{\vcat}[1]{\rotatebox[origin=c]{90}{\textbf{#1}}}
\begin{table}[htb]
\caption{Challenges implementing FAIR principles~\cite{wilkinson2016fair}}
\label{tab:gofair-requirements}
\renewcommand{\arraystretch}{1.2}
\begin{tabularx}{\linewidth}{lcp{5.5cm}c}
\toprule
\textbf{} & \textbf{ID} & \textbf{Description} & \textbf{Challenge}  \\
\midrule

\multirow{4}{*}{\centering \vcat{Findable}} 

& \Fone & (Meta)data are assigned a globally unique and persistent identifier 
& C3
\\

& \Ftwo & Data are described with rich metadata (defined by \Rone ~below) 
& C3
\\ 

& \Fthree & Metadata clearly and explicitly include the identifier of the data they describe 
& C3
\\

& \Ffour & (Meta)data are registered or indexed in a searchable resource 
& C5
\\

\midrule

\multirow{4}{*}{\centering \vcat{Accessible}} 

& \Aone & (Meta)data are retrievable by their identifier using a standardized communication protocol 
& C5
\\

& \Aoneone & The protocol is open, free, and universally implementable 
& C5
\\

& \Aonetwo & The protocol allows for authentication and authorization, where necessary 
& C5
\\

& \Atwo & Metadata remain accessible even when data are no longer available 
& C5
\\
\midrule
\multirow{3}{*}{\centering \vcat{Interoperable}} 
& \Ione & (Meta)data use a formal, accessible, shared, and broadly applicable language for knowledge representation 
& C1
\\
& \Itwo & (Meta)data use vocabularies that follow FAIR principles 
& C1
\\
& \Ithree & (Meta)data include qualified references to other (meta)data 
& C1
\\

\midrule
\multirow{4}{*}{\centering \vcat{Reusable}} 

& \Rone & (Meta)data are richly described with accurate and relevant attributes 
& C1, C3
\\

& \Roneone & (Meta)data are released with a clear and accessible data usage license 
& C5
\\

& \Ronetwo & (Meta)data are associated with detailed provenance 
& C4
\\

& \Ronethree & (Meta)data meet domain-relevant community standards 
& C2
\\

\bottomrule
\end{tabularx}
\vspace{-1.25\baselineskip}
\end{table}

Two challenges require (partial) support of tooling to manage all the (meta)data involved in a robotics pipeline, especially as it scales.
\textbf{C3: Automatic collection of (meta)data}. 
Given the scale of robotics datasets, the (meta)data collection and processing must be automated. 
This requires changes to tooling that transforms or generates artifacts to collect (meta)data and assign globally unique and persistent identifiers (\Fone) to relevant (meta)data for reuse  and include rich descriptions (\Ftwo, \Rone).
\textbf{C4: Automatic collection of provenance}. Similar to the collection of the (meta)data, provenance (\Ronetwo) requires automation, and thus supporting tools. 
For both \textbf{C3} and \textbf{C4}, implementation must consider that
robotics software is composed of many components. 
Some (static) components do not enable developers to modify the way they create output data, so metadata collection must be done via a "wrapper".
Furthermore, some metadata can be implicit (e.g., default parameter values) and needs to be identified and extracted, or only temporarily available (e.g., on real-world robots some metadata may only be available at startup).
Finally, huge data-size and fast streaming may require specialized implementations to record only relevant data (e.g., value changes) 
and for multi-robot datasets the ability to separate data originating from different agents.

Finally, \textbf{C5: Publishing robotics datasets} relates to creating the publishable dataset. Dataset publication requires a searchable index where metadata can be registered (\Ffour), which meets the requirements \Aone-\Atwo, and the choice of a license suitable for a variety of data (\Roneone).
Practical challenges are related to packaging datasets into subsets of files due to large sizes of the data (e.g., sensor data, video).

\section{Modeling provenance}
\label{sec:modelling}

\noindent
We now describe the models and metamodels used to represent the files, tools, activities, and agents involved in dataset creation. For knowledge representation, we use JSON-LD\footnote{\url{https://json-ld.org/}}, which supports independently defined domain-specific models that can later be composed into a single queryable graph. This machine-readable model consistently links campaign entities, their relationships, and provenance across all workflow steps. Because JSON-LD can be interpreted as RDF triplets, it also enables querying the graph with SPARQL, as shown in~\cref{sec:dataset}.

\subsection{Metamodels}

\noindent
We use well-established vocabularies to model the relationships between the activities and the elements in the campaign: 
PROV-O, DCAT and DCTERMS. 
To handle environment model metadata, we use the suggested metadata in~\cite{ieee7300355}.
Due to a lack of other standardized vocabularies in robotics, additional metadata from other artifacts and models uses a custom vocabulary in the \textit{robovast/metamodels/} namespace.
The QUDT\footnote{\url{http://qudt.org/schema/qudt}} vocabulary is used to describe the units used in the concepts.
All used vocabularies are included as prefixes in the JSON-LD document context, and for the rest of the paper are referred to as \textit{prov:}, \textit{dcat:} and \textit{dct:}, respectively.

\begin{figure}
    \centering
    \includegraphics[width=0.9\linewidth]{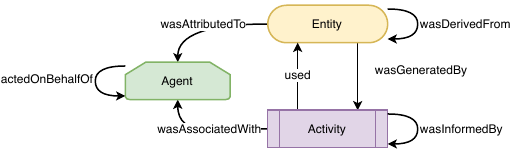}
    \caption{PROV-O concepts. Metamodels in this paper use the same color-coding scheme for entities, agents and activities.}
    \label{fig:prov-o-concepts}
\vspace{-0.5\baselineskip}
\end{figure}

The PROV standard is the core of our metamodel.
A metamodel is a model that describes how other models should be constructed.
In our case, PROV concepts and relationships based on the PROV-O ontology, shown in~\cref{fig:prov-o-concepts}, are used as the base representation for all elements in our dataset.
At its core, PROV models the generation, derivation, usage and attribution of agents (e.g., a robot), entities (e.g., a dataset or video file) and activities (e.g., a test run).
We use DCAT to model the dataset itself and other (meta)data entities. 
A \emph{distribution} represents a specific serialization of a dataset, e.g., zip files containing different groups of files.
Dublin Core terms are used where possible to model the metadata of the dataset contents. 

To create our metamodel, we examine the dataset creation process to identify entities, agents and activities.
Examples of entities include abstract scenarios (parametrized scenario families), concrete scenarios (fully instantiated test cases), environments (FloorPlan models), and robot configurations. 
Activities---such as generation or transformation of models or scenarios---link these entities with agents (e.g., persons, software agents, or the robot).
The metamodel provides a shared conceptual foundation established during design; however, the metamodels and choice of metadata evolve through an iterative feedback loop, with misalignments and inconsistencies identified during execution informing refinements to both the models and the tooling.

\begin{figure}
    \centering
    \includegraphics[width=0.85\linewidth]{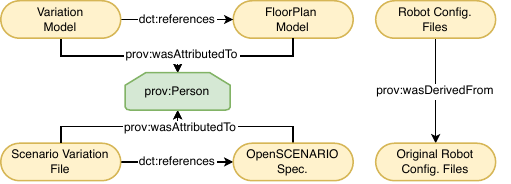}
    \caption{Provenance of the scenario inputs}
    \label{fig:prov-scenario-inputs}
    \vspace{-1.25\baselineskip}
\end{figure}

The campaign execution depends on a structured set of input artifacts
that collectively define the test scenarios, environments, and system configurations. 
These inputs are not isolated files but are connected through a network of explicit dependencies and metadata that track their origins, versions, and transformations.
The \textit{dct:references} relation links variations back to their source models and referenced input files, making the dependency graph explicit. The \textit{prov:wasAttributedTo} relation attributes the models and variations to their creators who are modeled as \textit{prov:Person}, ensuring authorship accountability. The \textit{prov:derivedFrom} relation links robot configuration files to their original counterparts, recording whether changes to the configuration were made. 
The \textit{dct:hasVersion} property 
captures version tags on each configuration artifact if available, and the  \textit{dct:modified} property records modification timestamps. 
The relations shown in~\cref{fig:prov-scenario-inputs} enable us to understand how input artifacts relate to one another and to trace each scenario instance back to the configuration sources that defined it.

\begin{figure*}
    \centering
    \includegraphics[width=0.85\linewidth]{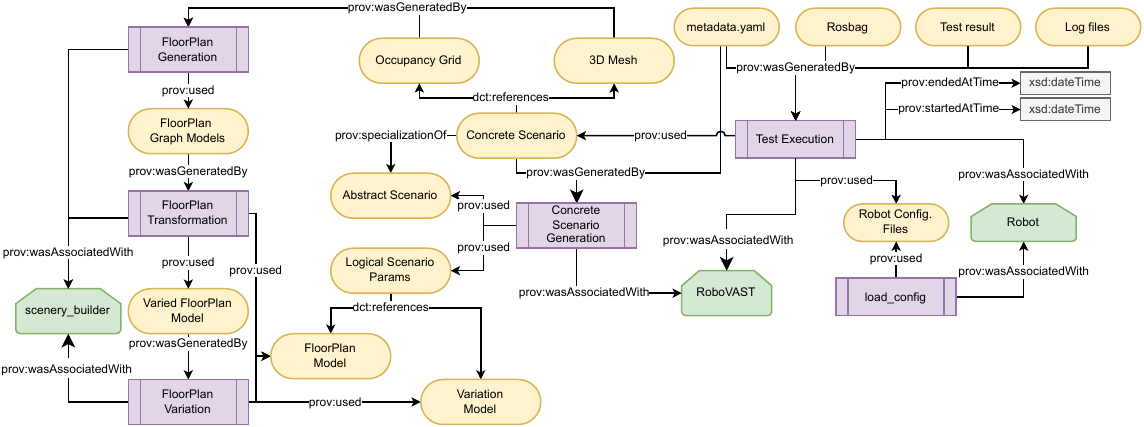}
    \caption{Provenance of scenario execution artifacts and artifact generation from Scenario Variation files}
    \label{fig:prov-scenario-gen-artifacts}
    \vspace{-1.25\baselineskip}
\end{figure*}

\cref{fig:prov-scenario-gen-artifacts} shows the metamodel for the scenario generation activities and artifacts.
On the left, three generation activities related to the environment models show the provenance relationships between artifacts (e.g., occupancy grid) used during the execution that can be used to trace back which models they were derived from (e.g., FloorPlan models).
These three activities are influenced by the generation activity on the right, which uses the OpenSCENARIO specification and the scenario variation file to generate the concrete scenario that will be executed.

The test execution activity generates some artifacts (e.g., bag file) while using as inputs the concrete scenario and the robot configuration.
The robot executing the test is modeled as an agent that is associated with the test execution activity.
To explicitly model the configuration the robot used for that run, we include a \emph{load\_config} activity that uses the configuration and is associated with the robot.
This chain of relationships between the entities in our dataset can be queried to identify, e.g., which input files were used for a given scenario, as will be shown in~\cref{sec:dataset}.

\subsection{Metadata}

\looseness-1
\noindent
In addition to provenance, we also model the properties of relevant elements in the test campaign.
Metadata captures both structural properties of dataset artifacts and domain-specific characteristics of validation campaign elements.
Metadata is captured at three levels: campaign, configuration, and test-run level.
These distinctions are necessary to differentiate between the execution context of a campaign and the properties of the scenarios within it. For example, the same configuration can be used in different executions 
with different start and end dates, software versions, and aggregate outcomes.
FAIR principles recommend adding as much metadata as possible to enable richer descriptions,
however, considering the modeling effort, we first prioritized metadata that would enable us to make such distinctions via graph queries.

We further enrich the campaign metadata by defining a metamodel with robot and framework-specific metadata using the \textit{robovast:} prefix. 
From the scenario variation parameters we include all the parameters that are used to create scenario variations and the number of runs in the execution parameters of the campaign.
The metadata of the environment models conforms to~\cite{ieee7300355}.
In the concrete scenarios, we record the number of runs and obstacles.
For the scenario execution, we include metadata that describes the system information, the metadata from the bag files generated by each run (e.g., ROS version and message types) and the result of the test run (success/failure).
Finally, for postprocessing steps we record the plugins and parameters that were used in the derivation of artifacts.

\subsection{Persistent identifiers in models}

\looseness-1
\noindent
We use Persistent URLs\footnote{\url{https://purl.archive.org}} (PURLs) as the node identifiers in the JSON-LD model for all elements in the dataset, with the following conventions. 
The base IRI for the Dataset is added in the model \textit{$@$context} using the \textit{$@$base} term. 
It uses PURL to make the IDs globally unique, persistent, and resolvable, and can be updated to point to the new address if the location ever changes.
The dataset entity also includes the \textit{dct:identifier} property whose value is the Zenodo DOI.
Each abstract scenario, concrete instance, execution, environment, and result is assigned an identifier based on their relative file path to the campaign results folder and using the base IRI of the dataset. 
Software agents also use PURL-based IRIs.
Finally, for human attribution, we use author's and contributors ORCID identifiers as their node \textit{$@$id}.

\section{Implementation}
\label{sec:implementation}

\noindent
To integrate the PROV and metadata collection into our framework, we follow an iterative process.
Pilot executions are used to identify missing metadata, and the schema and collectors are refined before large-scale campaign runs.

\subsection{Campaign Preparation}

\looseness-1
\noindent
As a first step, we manually add missing metadata to manually created files.
For example, for each environment model (\textit{.fpm}), we add a \textit{metadata} block that records the attribution, timestamps, size, authors, license, and a short human-readable description and map location, conforming to the metadata described in~\cite{ieee7300355}. 
Similarly, for each scenario variation description (\textit{.vast}), we add metadata about the involved agents and their configuration files, listing, for instance, which launch and parameter files are used by the navigation stack and which model and launch files compose the simulation setup.
We also include the necessary metadata to create the dataset models using DCAT, as described in~\cref{sec:modelling}.
The PURL of the dataset which is used as the base IRI for all other elements in the dataset is also specified in this section as can be seen in~\cref{lst:vast-metadata}.

\begin{lstlisting}[caption=Subset of manually specified metadata for a dataset, label=lst:vast-metadata]
  metadata:
    title: Navigation Dataset
    description: A navigation dataset that...
    creators:
      - ...
    keywords: ["robotics", "navigation", "ROS2"]
    license: "CC-BY-4.0"
    dataset_iri: https://purl.org/...
\end{lstlisting}
\vspace{-0.5em}

To automate the metadata extraction, we re-designed our framework to allow developers to integrate metadata collection plugins.
Using these newly developed plugins, \RoboVAST\ collects framework-specific metadata across multiple levels of the validation hierarchy.
Campaign-level metadata describes a specific execution effort (when and how a batch was run), including the number of planned runs, execution timestamps, and campaign-wide configuration. Structural metadata links abstract scenarios 
to their concrete instantiations, 
including which variations define the campaign and how they map to specific configurations.
Configuration-level metadata describes the reusable scenario definition itself (what was run), including: 
Nav2 parameters, %
environment references (FloorPlan models, 3D mesh and occupancy grids); 
and scenario parameters (e.g., number of goal poses and obstacle poses). 
Test run metadata documents execution outcomes by parsing \textit{test.xml} files to extract test success/failure status, execution duration, and precise start/end timestamps. 
System information includes hardware specifications, ROS distribution version, and runtime environment details. Output artifact metadata catalogs all generated files (rosbag recordings, log data) with their relative paths, while postprocessing metadata documents any transformations applied to raw outputs.

We extended \RoboVAST\ to also create and publish the distributions chosen in the design phase.
\cref{lst:vast-publication} shows a small example of how users can specify the publication of a distribution.
The inclusion filter allows developers to group dataset contents into meaningful groups (e.g., metadata, raw data, video, etc.).
The publication plugin takes care of the compression, and automates the upload of the distributions to Zenodo via its REST API.

\begin{lstlisting}[caption=Publishing specification in our framework, label=lst:vast-publication]
  publication:
  - zip:
      filename: "{timestamp:%Y-%m-%d}-graph.zip"
      include_filter:
      - "*.json"
\end{lstlisting}

    \vspace{-0.85\baselineskip}

\subsection{Campaign Execution}

\looseness-1
\noindent
The execution begins by transforming the abstract scenario specifications and variation templates into fully concrete, instantiated scenario instances. The scenario variation file, together with the \RoboVAST\ framework and input models (cf.~\cref{fig:prov-scenario-inputs}) drives this transformation.
For each concrete instance, the framework generates a complete set of resolved artifacts: an instantiated scenario configuration (\textit{scenario.config}), an occupancy grid and mesh representation of the environment, and associated metadata. Each instantiation resolves all parameterized values—environment layout, robot start pose, goal sequence, obstacle placements, sensor configurations, and random seeds—so that nothing is left implicit or underspecified. This resolution is the critical step that makes the scenario repeatable: the instantiated scenario and generated environment files form a self-contained, stand-alone specification of the scenario for future reuse. 

Test execution, shown on the right side of~\cref{fig:prov-scenario-gen-artifacts}, runs each instantiated scenario using \RoboVAST\ and automatically records the provenance and metadata produced in this stage.
It associates the test execution activity to a robot instance, its software parameters and the concrete scenario that it executes.
As the test run progresses, the framework captures: 
(1) ROS topics in a bag file; %
(2) execution metadata including start and end times, and hardware/software system details; 
(3) test results (pass/fail) from the Nav2 action. 
Upon completion of each test run, a \textit{test.xml} file records the test result, a \textit{metadata.yaml} file captures execution context, and log files document any runtime issues or diagnostic messages. Each test run receives a deterministic identifier derived from its standardized run-directory path, ensuring that provenance records are consistent for re-execution.

\looseness-1
The resulting artifacts are then organized so that runs, scenarios, and configurations can be understood and located without additional tooling. All artifacts collected during execution are exported into a standardized directory structure organized hierarchically across campaign-level configurations, scenario definitions, and individual execution runs. This organization ensures that all generated data, including scenario and environment configurations, are stored alongside execution outputs as a coherent dataset.
Provenance relationships are recorded automatically, linking each artifact to the concrete scenario and robot configuration that produced it (with execution timestamps), enabling third parties to trace which inputs were used for each test and reconstruct the conditions under which evidence was generated shown in~\cref{fig:prov-scenario-gen-artifacts}.

\subsection{Dataset Publication and Distribution}

\looseness=-1
\noindent
This phase postprocess raw execution outputs into easily analyzed formats and publishes them as structured distributions with comprehensive metadata and provenance.
Raw bag files produced during execution are not immediately analysis-ready: they require specialized ROS tools to access, bundle hundreds of message topics (many irrelevant to navigation analysis), and encode data in binary formats. To enable broader reuse and simplify analysis workflows, the post-processing phase derives two categories of standardized outputs from each raw execution. First, navigation-relevant time series (e.g., pose estimates)
are extracted into CSV files, providing tabular data accessible to any analysis tool without ROS dependencies. Second, videos are generated from Gazebo camera frames, enabling visual inspection of robot behavior and environment interactions without rerunning simulations. 
Derived artifacts enable analysis by standard tools (e.g., Python)
without requiring ROS.

After postprocessing raw data, \RoboVAST\ automatically orchestrates metadata and provenance consolidation from different stages. %
That is, it processes all the metadata collected throughout the campaign activities: input dependencies, scenario generation, test execution, and post-processing transformations.
For each test run, the pipeline parses scenario configuration files, extracts test outcomes and system configuration, and catalogs input files and output artifacts with their paths and identifiers.

Next, we convert the collected metadata into a linked graph.
This process is done by matching the relevant metadata with the metamodels described in~\cref{sec:modelling}, so that concepts have an \textit{$@$id} relative to the dataset base IRI, have the correct \textit{$@$type} and any other relevant properties included in our metamodel.
The metamodel describes which of the properties are interpreted as IRIs, and thus establishes relationships between items in the dataset.
Because provenance metadata is published as JSON-LD with explicit semantic relations, researchers can query the dataset structure to answer questions about campaign composition, scenario configurations, and artifact dependencies.

Publication and archival turns the validated and processed dataset from an internal resource into a publicly discoverable, permanently archived research asset. 
By packaging (meta)data into distinct distributions and publishing through institutional repositories, the dataset becomes discoverable, citable, and accessible for reuse. %

\section{Our FAIR Dataset}
\label{sec:dataset}

\begin{figure}[tbp]
    \centering
    \includegraphics[width=\linewidth]{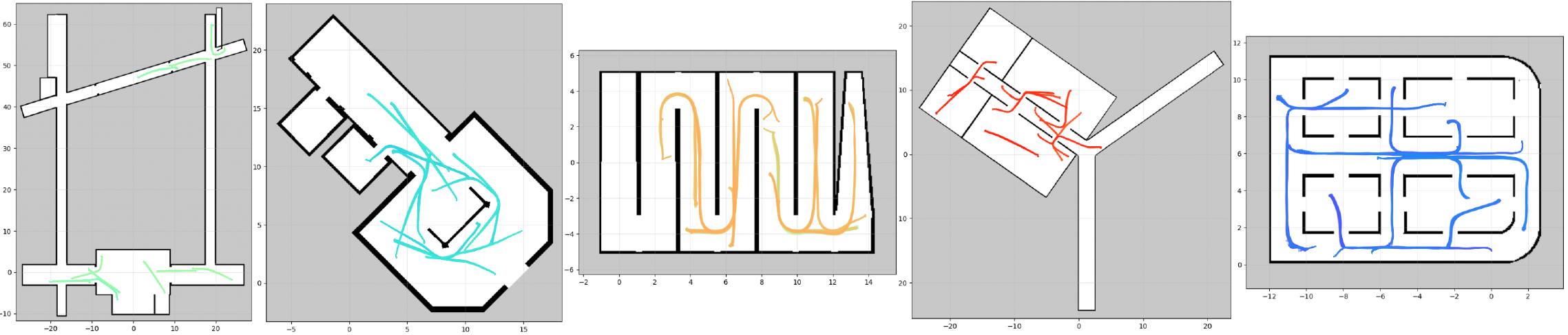}
    \caption{Maps employed in the dataset with traversed paths.}
    \label{fig:dataset-contents}
\end{figure}

\noindent
Our validation dataset is focused on scenario testing.
The scenario inputs consist of the SUT introduced in~\cref{sec:case-study} and the indoor maps shown in~\cref{fig:dataset-contents}. We define \num{400} configurations, \num{10} runs each,  over \num{5} maps generated with \textit{FloorPlanGeneration}. The leftmost is based on a real university floorplan~\cite{Parra2023}, while the others are synthetic. For each map, we use 10 unique \qty{10}{\metre} paths created by \textit{PathVariationRandom}. Static obstacles are added with \textit{ObstacleVariation} at densities of \{\num{0}, \num{0,2}\} obstacles per meter of path length, and the Nav2 local costmap robot radius is varied via \textit{ParameterVariationList} with values of \qty{0.175}{\metre} and \qty{0.22}{\metre}.

\looseness-1
Our campaign results in \num{4000} runs, with \num{290} failed runs (\num{7.22}\% failure rate). Across all runs, the robot covered a cumulative ground-truth distance of \qty{39.3}{\kilo\metre}. On the right-most map, most failures occurred at startup where spawning in the top left corner always caused start up errors due to the robot spawning within the inflation radius of the wall in \num{8} configs, leading to \num{80} failed runs. Also of note are two configurations on the second map from the right where the robot consistently (\num{19} out of \num{20} runs) ran into an obstacle at the intersection, even given sufficient room to pass. 
The dataset graph consists of \num{307127} triples.
We summarize our compliance to FAIR principles in~\cref{tab:fair-results}.
The dataset and code are published\footnote{\url{https://doi.org/10.5281/zenodo.18702398}} under permissive open-access licenses.

\setlength{\tabcolsep}{2pt}
\begin{table}[tbp]
    \centering
    \caption{Compliance with FAIR principles. Full and partial compliance are marked with green and orange, respectively. Red indicates non-compliance and open challenges.}
    \label{tab:fair-results}
    \begin{tabular}{lp{0.83\linewidth}}
    \toprule
        Principle & Means of achievement\\
        \midrule
         \Fone-\Ffour & Distributions include rich metadata (\Ftwo \tl{green}) with persistent, globally unique identifiers  (DOIs, ORCID and PURL) (\Fone \tl{green} \Fthree \tl{green}). Publishing in Zenodo. (\Ffour \tl{green}).  \\
         \Aone-\Atwo & Zenodo is accessible via HTTP and has a REST API (\Aone \tl{green}); it provides long-term data preservation (20-year commitment) (\Atwo \tl{green}).  Data is indexed and searchable.\\
         \Ione-\Ithree & Modeling using well-known vocabularies: PROV, DCAT, Dublin Core and QUDT. JSON-LD allows the use of RDF tools and SPARQL queries (\Ione \tl{green}). However, we use custom vocabularies for robotics- and framework metadata (\Itwo \tl{orange}). Our models do not always use qualified references (\Ithree \tl{orange}).\\
         \Rone--\Ronethree & Dataset and code have permissive open-access licenses (\Roneone \tl{green}). We explicitly model provenance of activities, entities and agents in our process (\Ronetwo \tl{green}). We use~\cite{ieee7300355} for environment metadata and include robotics and framework-specific metadata (\Rone \tl{orange}). However, lack of community standards and controlled vocabularies remain open challenges for \Ronethree \tl{red}\\
         \bottomrule
    \end{tabular}
    \vspace{-1.5\baselineskip}
\end{table}

The linked graph supports replicability through queries to, for example, retrieve the input files, the task parameters and the results of each scenario in the dataset.
\Cref{lst:replicability-query} queries all the input files for all the executions and groups them by concrete scenario, listing all the required input files to replicate the results in our dataset. 
The start pose and navigation goals are stored in the \textit{ConcreteScenario} configuration file returned by this query.
The last two lines in~\cref{lst:replicability-query} limit the results to scenarios without obstacles, in case the user wants to replicate scenarios without using \RoboVAST. 

\begin{lstlisting}[caption=Simplified query for the inputs of the test executions for a concrete scenario, xleftmargin=1.5em, label=lst:replicability-query,language=SPARQL]
SELECT ?conf
GROUP_CONCAT(DISTINCT ?f;SEPARATOR=",") AS ?fs
WHERE {
  ?run rdf:type robovast:TestExecution .
  ?conf rdf:type smm:ConcreteScenario .
  ?run prov:used/(dcterms:references|prov:hadMember|prov:atLocation)* ?f .
  ?conf robovast:n_obstacles ?obst .
  FILTER ( ?obst = 0 )
} GROUP BY ?conf
\end{lstlisting}
\vspace{-0.5em}

Finally, \cref{lst:eval-query} shows a query for retrieving the failure rate of each concrete scenario, enabling users to access and summarize results for the dataset parts they want to replicate. Together with \cref{lst:replicability-query}, it demonstrates how the provenance relationships in~\cref{fig:prov-scenario-gen-artifacts} preserve traceability between entities used or generated during campaign execution and how it can be accessed by other users after the fact.

\begin{lstlisting}[caption=Query the failure rate per concrete scenario, label=lst:eval-query,language=SPARQL]
SELECT ?conf (SUM(?fail)/COUNT(?run)*100 AS ?rate) (COUNT (?run) AS ?total)
WHERE {
  ?conf rdf:type smm:ConcreteScenario .
  ?run prov:used ?conf .
  ?run rdf:type robovast:TestExecution .
  ?run robovast:success ?success .
  BIND(IF(?success=true, 0, 1) AS ?fail) .
} GROUP BY ?conf
\end{lstlisting}
\vspace{-1.2em}

\section{Conclusions}

\noindent
Simulation-based validation is widely used to validate robot behavior and have great potential for replication studies.
We showed how provenance models document the traceability between files, tools, agents and processes involved in a validation campaign, and how queries to the provenance linked graph provide richer, machine-readable descriptions of the (meta)data in the generated dataset.
Although our robotics- and framework-specific metamodels, as well as our implementation, may be hard to generalize, our PROV metamodel gives a much richer description of what is included in our dataset and under which conditions it was executed.
Compliance with FAIR principles requires further work and community involvement, as
it requires substantial upfront investment
which may be infeasible for smaller groups or time-constrained campaigns. 

Future work includes generalizing reusable components, 
contributing vocabulary extensions to standardization efforts, and developing lightweight tools for incremental FAIR compliance verification. 
For modeling, we plan on generalizing concepts, including other scenario and robot types and improving composition in our (meta)models, adding qualified references and implementing model validation across metadata and provenance graphs.
An advantage of the composability of JSON-LD documents is that the provenance recording can be decentralized and modular, 
allowing other provenance recording tools to be developed independently and composing their resulting graphs with our models.

\bibliographystyle{IEEEtran}
\bibliography{bibliography}

\end{document}